\documentclass[runningheads]{llncs}

\usepackage{eccv}

\usepackage{eccvabbrv}

\usepackage{graphicx}
\usepackage{booktabs}

\usepackage[accsupp]{axessibility}  

\usepackage[pagebackref,breaklinks,colorlinks]{hyperref}

\def\mc{\mathcal}

\usepackage{multirow}
\usepackage{graphicx}
\usepackage{caption}
\usepackage{booktabs}
\usepackage{xcolor}
\def\du{\textcolor{red}}

\usepackage[marginal]{footmisc}

\begin{document}


\title{Confidence Self-Calibration for Multi-Label Class-Incremental Learning}


\author{Kaile Du\inst{1} \and
Yifan Zhou\inst{1} \and
Fan Lyu\inst{2} \and
Yuyang Li\inst{1} \and
Chen Lu\inst{1} \and
Guangcan Liu\inst{1}
}

\authorrunning{K.Du, Y.Zhou et al.}

\institute{School of Automation, Southeast University, China \and
CRIPAC, MAIS, CASIA, China
}

\maketitle

\begin{abstract}
The partial label challenge in Multi-Label Class-Incremental Learning (MLCIL) arises when only the new classes are labeled during training, while past and future labels remain unavailable. 
This issue leads to a proliferation of false-positive errors due to erroneously high confidence multi-label predictions, exacerbating catastrophic forgetting within the disjoint label space.
In this paper, we aim to refine multi-label confidence calibration in MLCIL and propose a Confidence Self-Calibration (CSC) approach.
Firstly, for label relationship calibration, we introduce a class-incremental graph convolutional network that bridges the isolated label spaces 
by constructing learnable, dynamically extended label relationship graph.
Then, for confidence calibration, we present a max-entropy regularization for each multi-label increment, facilitating confidence self-calibration through the penalization of over-confident output distributions. 
Our approach attains new state-of-the-art results in MLCIL tasks on both MS-COCO and PASCAL VOC datasets, with the calibration of label confidences confirmed through our methodology.
\end{abstract}

\footnote{Kaile Du and Yifan Zhou are co-first authors, \email{kailedu@seu.edu.cn}. Correspondence to: Guangcan Liu, \email{gcliu1982@gmail.com}.}
\section{Introduction}
\label{sec:intro}
\begin{figure}[t]
    \centering
\includegraphics[width=\linewidth]{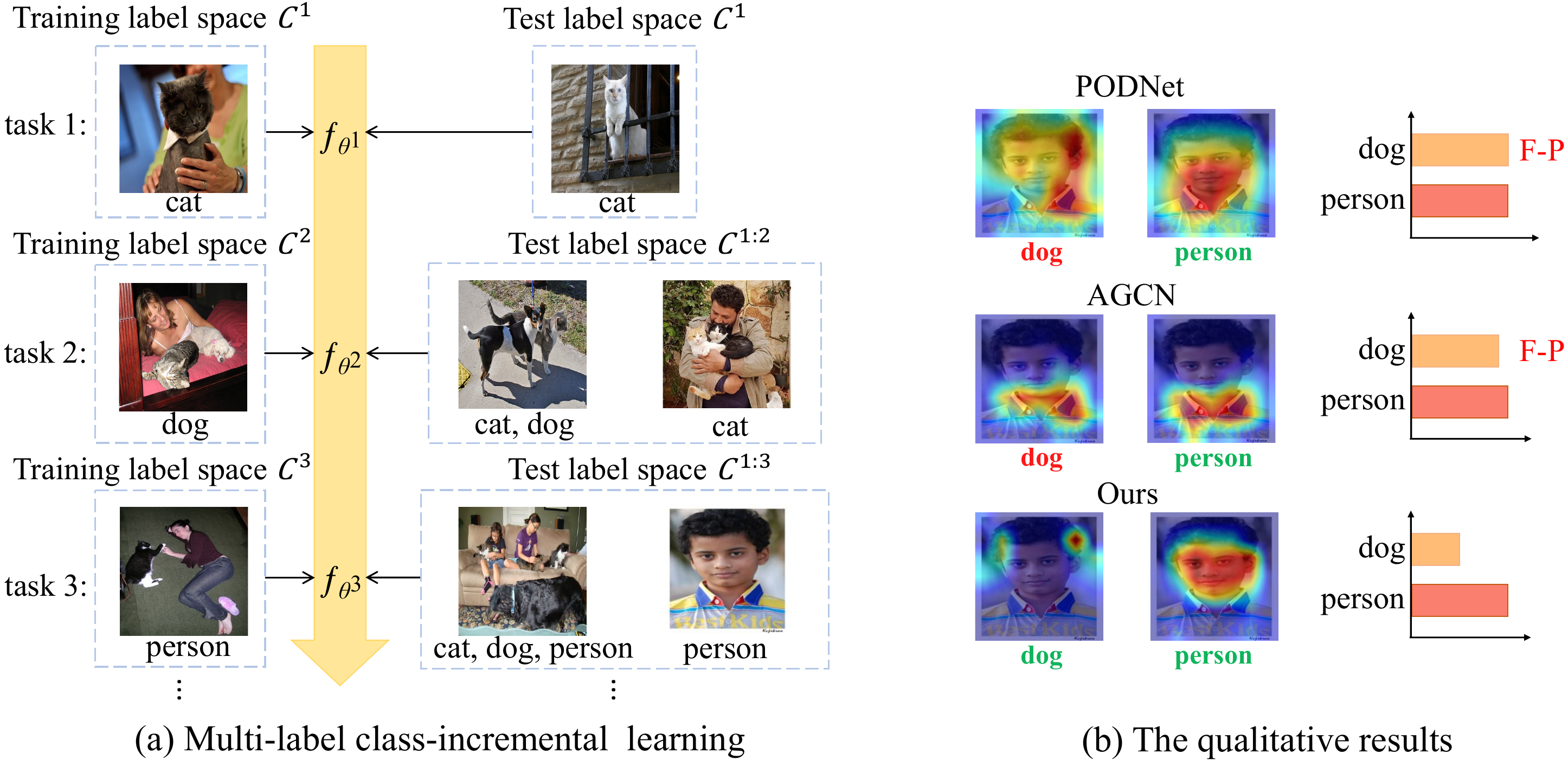}
    \caption{ (a) A simple depiction of multi-label class-incremental learning. There are a total of three labels, including ``cat'', ``dog'' and ``person'', which are sequentially learned and tested across three tasks. (b) After training on task 3, given a test image only with ``person'' processing by $f_{\theta^3}$. Left: attention maps for the labels ``dog'' and ``person'' using the SOTA methods PODNet \cite{douillard2020podnet}, AGCN \cite{10221710} and ours. Right: The confident scores for the three methods. The false-positive (\du{F-P}) error occurs on the ``dog'' using PODNet and AGCN. The bottom shows the effectiveness of our method. 
    }
    \label{fig:mlcil}
\end{figure}

Class-incremental learning (CIL) \cite{zhou2022few, mai2021supervised,10221710, wang2023comprehensive,lyu2024elastic,lyu2024variational} refers to learning a sequence of new classes and being able to preserve the knowledge learned well. 
{In CIL, the model learns only from samples associated with the incremental classes, leading to the challenge of catastrophic forgetting} \cite{mccloskey1989catastrophic}, 
where adaptation to new classes often results in a significant decline in performance on the old ones.
Most CIL methods assume single-label annotations for inputs and employ techniques to mitigate single-label aware forgetting \cite{buzzega2020dark,lyu2023measuring,sun2022exploring,li2023long}. 
In contrast, there is limited exploration of 
multi-label class-incremental learning (MLCIL), which aims to continually adapt to the increasingly larger label space through non-forgetting learning.
MLCIL focuses solely on learning the incremental classes, resulting in the absence of past and future labels, a phenomenon referred to as the \textit{task-level partial label}  \cite{10221710} or category-incomplete \cite{dong2023knowledge} issue.
A toy-like MLCIL depiction is shown in  \cref{fig:mlcil} (a). Given three labels in total, including ``cat'', ``dog'' and ``person'', which are sequentially learned and tested across three tasks. During task 2, only the current task class (``dog'') is labeled at training time, while the classes from the previous (``cat'') and future (``person'') tasks remain unlabeled. 

Partial label issue makes the new and old label spaces naturally disjoint. The previous efforts \cite{dong2023knowledge, 10221710}
have overlooked the need to calibrate the MLCIL model, often resulting in an over-confident output distribution \cite{pereyra2017regularizing,rizve2021defense} with false-positive (F-P) errors \cite{song2021handling}. 
As shown in \cref{fig:mlcil} (b), 
even when presented with a test image containing only a person, SOTA methods like PODNet \cite{douillard2020podnet} and AGCN \cite{10221710} still output high confidence for both old class ``dog'' and new class ``person''. 
This phenomenon indicates an F-P error 
and becomes rigorous with the enlargement of the label space of MLCIL. 
The erroneous over-confident multi-label predictions from poorly calibrated models will degrade performance on old classes, exacerbating catastrophic forgetting under the disjoint label space.

  In this paper, we aim to calibrate MLCIL model and propose a Confidence Self-Calibration (CSC) approach under the partial label setting, comprising two components: 
(1) a class-incremental graph convolutional network (CI-GCN) structure for label relationship calibration
  and 
  (2) a max-entropy regularization for confidence self-calibration. 
   Firstly, while relationships between labels are beneficial in multi-label classification~\cite{zhao2021transformer, chen2021learning, ye2020attention}, with the partial label issue, it becomes challenging to construct cross-task relationships due to the missing labels from disjoint label space. To address this issue, we propose the CI-GCN to calibrate the label relationships. The label relationship graph constructed by CI-GCN is continuously learned through gradient updates. Under the partial label challenge, CI-GCN not only preserves label relationships in the old label space but also consistently expands to establish relationships in the new label space.
Secondly, miscalibrated confidences result in misclassifications in multi-label problems. To mitigate this issue, we propose a max-entropy regularization 
 to penalize over-confident output distribution, effectively reducing the F-P rate to facilitate confidence self-calibration.

Our main contributions include:
\begin{itemize}
    \item {A CI-GCN is designed for label relationship calibration, which bridges isolated label spaces by establishing cross-task label relationships. The dynamically extended label relationship graph is continually learned through gradient updates. }
    \item  A max-entropy regularization is proposed to facilitate the confidence self-calibration by penalizing over-confident output distribution, effectively reducing the false-positive errors.
    \item Extensive experimental results on MS-COCO and PASCAL VOC datasets demonstrate that the proposed method distinctly outperforms recent state-of-the-art approaches in MLCIL. 
\end{itemize}

\section{Related Work}

\noindent
\textbf{Single-Label Class-Incremental Learning.} 
 Recent advances have been made in single-label class-incremental learning. For example, the regularization-based methods \cite{kirkpatrick2017overcoming,li2017learning,zhou2022few,zhao2023few,mohamed2023d3former,9859622} incorporate a regularization term into the loss function to learn new classes without forgetting the old ones. EWC \cite{kirkpatrick2017overcoming} utilizes a Fisher matrix to retain critical parameters from previous tasks. Then, oEWC \cite{schwarz2018progress} proposes an improvement to the calculation of parameter importance.
The replay-based methods \cite{rebuffi2017icarl,cha2023rebalancing,ye2021lifelong,de2021continual,rolnick2019experience,shin2017continual,luo2023class,liu2023centroid,lyu2021multi} overcome forgetting by storing partial data of learned classes in a small buffer and employing suitable sampling techniques. ER \cite{rolnick2019experience} trains with samples from old tasks in the buffer alongside new task samples. 
Building upon this, DER++ \cite{buzzega2020dark} integrates rehearsal with knowledge distillation and regularization, while iCaRL \cite{rebuffi2017icarl} prevents forgetting by selecting exemplars with the herding technique. And the architecture-based approaches, which decompose a model into task-sharing and task-specific components \cite{douillard2022dytox,wu2021incremental}, have also been proposed. Moreover, L2P \cite{wang2022learning} introduces prompt-based learning \cite{liu2023pre} into class-incremental learning and achieves promising results based on a pre-trained ViT model \cite{dosovitskiy2020image}.

\noindent
\textbf{Multi-Label Classification.} Multi-label classification has garnered widespread attention
 Relation-based methods\cite{chen2019learning,chen2019multi,wang2017multi,wang2016cnn,chen2020knowledge,chen2021learning} aim to leverage the label dependencies or region relations for image classification. Earlier approaches utilize RNNs \cite{wang2016cnn} or LSTMs \cite{wang2017multi} to model the relationship between labels and semantics sequentially, while state-of-the-art methods leverage spatial information or graphical architectures. 
  For example, Chen et al. \cite{chen2021learning} propose a multi-label classification model based on graph convolutional network to generate inter-dependent object classifiers. 
  KGGR \cite{chen2020knowledge} exploits prior knowledge to guide adaptive information propagation among  categories to facilitate multi-label analysis, 
  while Zhao et al.\cite{zhao2021transformer} construct complementary relationships by exploring structural and semantic correlation graphs. 
 Our work is inspired by the GCN-based methods.

 \noindent
\textbf{Multi-Label Class-Incremental Learning.} With the ongoing research in the fields of class-incremental learning and multi-label classification, multi-label class-incremental learning has emerged as a new area of interest. 
OCDM \cite{liang2022optimizing} improves the replay-based method by formulating the memory update mechanism as an optimization problem, while PRS \cite{kim2020imbalanced} proposes a novel sampling strategy for replay-based approaches to mitigate the impact of long-tail distribution in multi-label datasets. Dong et al.  \cite{dong2023knowledge} propose a knowledge restoration and transfer framework to address the issues of label absence.
AGCN \cite{10221710} utilizes pseudo labels to construct fixed statistical label relationships and mitigate catastrophic forgetting.  Note that KRT \cite{dong2023knowledge} is the current SOTA method.
However, these methods have failed to recognize the significant occurrence of F-P errors caused by over-confident output distribution, thus lacking in model calibration.

\section{Method}
\subsection{Problem Formulation}

 Following \cite{10221710,dong2023knowledge}, given \textit{T} incremental learning tasks, with the corresponding training set denoted as $\{\mc{D}^{1}_\text{trn},\cdots,\mc{D}^{T}_\text{trn}\}$ and the testing set as  
 $\{\mc{D}^{1}_\text{tst},\cdots,\mc{D}^{T}_\text{tst}\}$. For each incremental state \textit{t}, the training dataset is defined as ${\mc{D}^{t}_\text{trn}=\{(x^t_i,y^t_i)\}}$, where $x^t_i$ is the \textit{i}-th training sample, and $\textit{y}^t_i \in \mc{Y}^t_\text{trn}$ denotes the corresponding ground-truth labels. 
 The training label space $\mc{Y}^t_\text{trn}=\mc{C}^t$, and $\mc{C}^t$ denotes the current class collection, and the past class collection is represented by $\mc{C}^{1:t-1}=\bigcup_{i=1}^{t-1}\mc{C}^i$.   $\mc{C}^{1:t} = \mc{C}^{1:t-1}\cup\mc{C}^{t}$, and $\mc{C}^{1:t-1}\cap\mc{C}^t=\varnothing$. The testing label space $\mc{Y}^t_\text{tst}$ is equivalent to $\mc{C}^{1:t}$ for the \textit{t}-th task, formulated as $\mc{Y}^t_\text{tst}=\mc{C}^{1:t}$. 
Under the task-level partial label condition, we intend to calibrate the model based on the connected label space. Our goal is to establish relationships between new and old task labels to connect the label space, and then calibrate the over-confident output distribution on this basis.

  \begin{figure}
    \centering
    \includegraphics[width=0.94\linewidth]{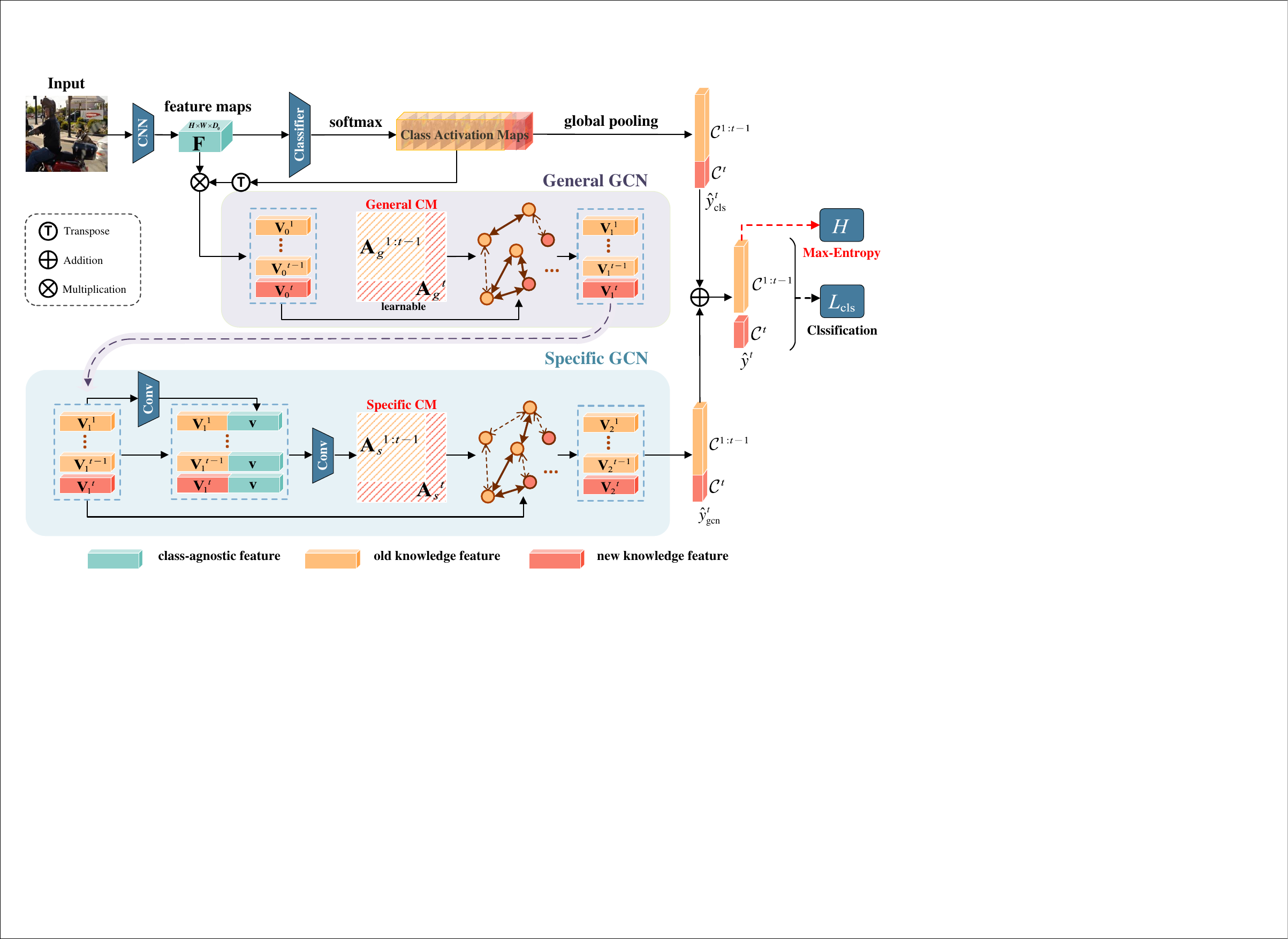}
    \caption{Overview of confidence self-calibration framework. Given an image, we first employ a CNN backbone to extract the class-agnostic feature maps \textbf{F}. Then, inspired by CAM \cite{zhou2016learning}, we use the class activation maps as masks to decouple \textbf{F} into class-aware label representations $\textbf{V}_0$. Our model constructs general and specific graphs among $\textbf{V}_0$ to generate $\textbf{V}_2$, which aggregates rich cross-task label relationships. Finally, we combine graph-based representations $\hat{y}^t_\text{gcn}$ with class-related outputs  $\hat{y}^t_\text{cls}$ for classification. 
    }
    \label{fig:CI-GCN}
\end{figure}
 
\subsection{Overall Framework}

We propose a Confidence Self-Calibration (CSC) framework under the partial label setting, comprising two major components: (1) A stacked \textbf{Class-Incremental Graph Convolutional Network} (CI-GCN) structure (\cref{sec:CI-GCN}), inspired by the GCN in traditional multi-label learning  \cite{zhao2021transformer, chen2021learning, ye2020attention}. 
CI-GCN first employs a feature extractor to obtain class-agnostic feature maps, followed by the stacked structure to generate cross-task graph representations for calibrating label relationships.
(2) A \textbf{Max-Entropy Regularization} (\cref{sec:max-ent}) applied to the old classes to calibrate multi-label confidence in MLCIL.
When CSC outputs predictions, it combines class-related features with graph representations to classify all the seen labels so far under the partial label condition.


\subsection{Class-Incremental Graph Convolutional Network}
\label{sec:CI-GCN}
Utilizing graph structure to construct label relationships has been proven effective in GCN-based methods~\cite{zhao2021transformer, chen2021learning, ye2020attention, chen2019multi}. 
These methods \cite{zhao2021transformer, chen2021learning, ye2020attention, chen2019multi, 10221710} rely on the input of graph nodes, \ie, multiple labels, and a correlation matrix (CM) to provide graph edges for label relationships.
However, these approaches necessitate prior statistical counting of label co-occurrences across the entire label set to construct the CM. Given the partial label setting, access to past and future labels is restricted, posing a challenge in constructing cross-task CM.
As shown in \cref{fig:CI-GCN}, CI-GCN is a stacked structure that does not depend on prior information, comprising one layer of general GCN and one layer of specific GCN. 
 A learnable general CM is input into the general GCN,  capturing general class dependencies by learning new and old task knowledge within the model through gradient updates during training, thus allowing the general GCN to automatically construct cross-task label relationships without relying on statistical relationships. Note that the general CM is shared for all test images in the same task during testing. 
Subsequently, the specific GCN is a finer structure to provide image-dependent relationships, where the specific CM generated by each image may vary. 
The CMs in the two GCNs will continually and automatically expand as the number of learned classes increases. 
General CM and specific CM capture general and specific class dependencies, respectively, thereby aiding in complementing missing label information from the overall to the detailed level.

\noindent
\textbf{Constructing CMs.} 
Before the first task, we randomly initialize the general CM $\textbf{A}_g^1$, which will expand and be updated in the subsequent tasks.  
A simple way to address partial supervision in MLCIL is to use pseudo labels from the previous model.
Inspired by the old and new knowledge tokens from KRT \cite{dong2023knowledge}, we divide the general CM into two parts shown in \cref{fig:CI-GCN}: the old task CM  $\textbf{A}_g^{1:t-1}$ is inherited from the previous tasks, preserving old task label relationships, and the new task CM $\textbf{A}_g^t$ establishes relationships in the new task label space. 
We employ the new supervision from the training set and the old supervision, \ie pseudo-labels, from previous model, to jointly update $\textbf{A}_g$ through gradient descent, with  $lr$ as the learning rate and $L$ representing the loss function, formulated as: 
\begin{equation}\label{eq:A_g}
\textbf{A}_g = \textbf{A}_g-lr*\frac{\partial L}{\partial \textbf{A}_g}, \quad \text{where~~}
\frac{\partial L}{\partial \textbf{A}_g} = 
\frac{\partial L}{\partial \textbf{V}_2} 
\frac{\partial \textbf{V}_2}{\partial \textbf{A}_g}=
\frac{\partial L}{\partial \textbf{V}_2} 
\frac{\partial \textbf{V}_2}{\partial \textbf{V}_1}
\frac{\partial \textbf{V}_1}{\partial \textbf{A}_g},
\end{equation}
enabling $\textbf{A}_g$ to learn cross-task label relationships, where $L$ takes both ground truth and pseudo-labels as input. $\textbf{V}_1$ and $\textbf{V}_2$ denote the outputs of the general and specific GCN shown in \cref{fig:CI-GCN}, respectively, which will be discussed next.

In previous GCN-based multi-label classifications \cite{zhao2021transformer, chen2021learning, ye2020attention, chen2019multi, 10221710}, the CMs are shared across all images, and directly counting pseudo-labels like AGCN \cite{10221710} will lead to significant error accumulation. Motivated by this, we design an adaptive CM that does not rely on pseudo-label statistics, providing unique relationships for each image without being significantly affected by error accumulation.
As shown in \cref{fig:CI-GCN}, after augmenting the general CM, the specific CM  $\textbf{A}_s$ is adaptively estimated from input graph nodes $\textbf{V}_1$, automatically expanding with the expansion of $\textbf{V}_1$.
As shown in \cref{fig:CI-GCN}, the specific CM is also divided into new and old parts, offering greater flexibility to adapt to continually changing CIL compared to fixed statistical CMs without compromising the performance of rare label combinations.
In detail, we begin by performing global pooling and applying a convolutional layer to $\textbf{V}_1$ to derive its global feature $\textbf{v} \in \mathbb{R}^{D_1}$.  
 This feature is then replicated  for $|\mc{C}^{1:t}|$ times and concatenated with $\textbf{V}_1$ shown in \cref{fig:CI-GCN}, yielding $\textbf{V}_1' \in \mathbb{R}^{|\mc{C}^{1:t}| \times 2D_1}$. Subsequently, we pass $\textbf{V}_1'$ through another convolutional layer with parameters $\textbf{W} \in \mathbb{R}^{2D_1 \times |\mc{C}^{1:t}|}$ to compute the sample-specific CM $\textbf{A}_s \in \mathbb{R}^{|\mc{C}^{1:t}| \times |\mc{C}^{1:t}|}$:
 \begin{equation}\label{eq:A_s}
      \textbf{A}_s = \sigma( \textbf{V}_1'\textbf{W}),
\end{equation}
where $\sigma$ is the $Sigmoid(\cdot)$ function. 

\noindent
\textbf{General and Specific GCN.} 
As shown in \cref{fig:CI-GCN}, the CI-GCN takes class-related label representations $\textbf{V}_0$ as graph nodes, calculated by:
 \begin{equation}\label{eq:v}
    \textbf{V}_0 = \textbf{M}^\top \otimes \textbf{F} ,
\end{equation}
 where $\textbf{F} \in \mathbb{R}^{H \times W \times D_0}$ is class-agnostic feature maps, and $\textbf{V}_0 \in \mathbb{R}^{|\mc{C}^{1:t}| \times D_0}$. 
 The class activation maps $\textbf{M} \in \mathbb{R}^{H \times W \times |\mc{C}^{1:t}|}$ are generated by the convolution classifier with $Softmax(\cdot)$ normalization shown in \cref{fig:CI-GCN}. 
Given the two CMs computed by Eqs.~\eqref{eq:A_g} and \eqref{eq:A_s}, the general and specific GCN can be defined as:
\begin{equation}\label{eq:gph2}
    \textbf{V}_1 = LReLU(\textbf{A}_g \textbf{V}_0\textbf{W}_g), ~~    \textbf{V}_2 = LReLU(\textbf{A}_s \textbf{V}_1\textbf{W}_s),
\end{equation}
where $LReLU(\cdot)$ represents the activation function LeakyReLU, which makes the whole operation nonlinear.
$\textbf{W}_g \in \mathbb{R}^{D_0 \times D_1}$ and $\textbf{W}_s \in \mathbb{R}^{D_1 \times D_2}$  denote the state-update parameters for general and specific GCN. $\textbf{V}_1 \in \mathbb{R}^{|\mc{C}^{1:t}| \times D_1}$  is the output of the general GCN, serving as the input notes for the specific GCN.  
To enhance model robustness, the specific GCN is employed to generate sample-specific label relationships for each image under the partial label condition. $\textbf{V}_2 \in \mathbb{R}^{|\mc{C}^{1:t}| \times D_2}$ represents the output of the specific GCN. 
{With the two GCNs, we complete the missing labels via two designed CMs.
However, as shown in Fig.~\ref{fig:ent}, we observe that the recall is generally much higher than precision, indicating that the confidence for each label is generally high, leading to many F-P errors. This phenomenon is also very evident in other CIL methods \cite{10221710, douillard2020podnet, buzzega2020dark,li2017learning,rolnick2019experience}.  We propose to calibrate this multi-label confidence via an entropy regularization.}
\subsection{Max-Entropy Regularization}
\label{sec:max-ent}
 \begin{figure*}[t]
    \centering
    \includegraphics[width=0.78\linewidth]{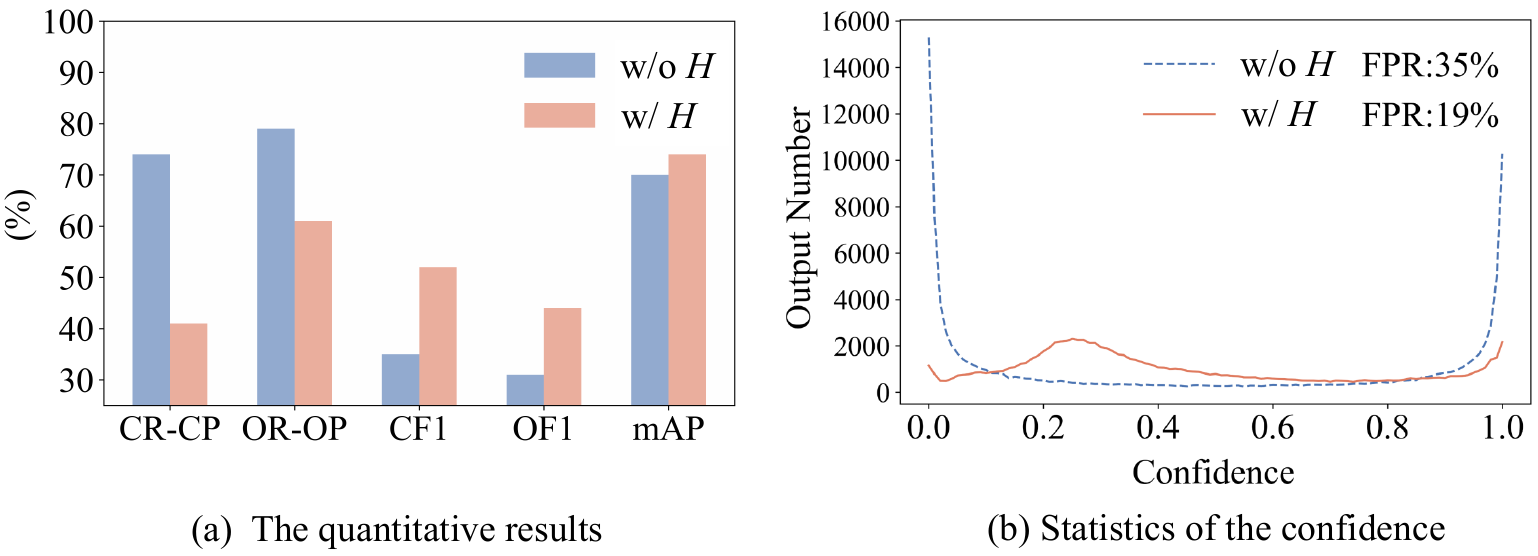}
    \caption{(a) The quantitative results without or with max-entropy based on CSC framework after training on the final task. CR-CP represents per-class recall minus precision and OR-OP represents overall recall minus precision. (b) The corresponding multi-label confidence statistics on the entire test set, as the information entropy $H$ increases from 1.61 (w/o \textit{H}) to 5.36 (w/ \textit{H}) in \{B4-C2\} of VOC 2007.
    }
    \label{fig:ent}
\end{figure*}
Building upon the calibrated label relationships, we proceed to calibrate multi-label confidence. 
Given that the current task classes are partially labeled, the model struggles to effectively discern between labeled and unlabeled class features. Consequently, the output distribution becomes over-confident, leading to numerous F-P errors.
 
 From the quantitative perspective:
 when incrementally training multiple labels, we observe a significant imbalance between the precision and recall values of the model (often with precision markedly lower than recall) shown in \cref{fig:ent} (a) CR-CP and OR-OP w/o \textit{H}. This discrepancy indicates a high frequency of F-P errors from an alternate perspective. 
  From the qualitative perspective: 
  we define P($l_\text{person}$|$l_\text{tie}$) as the conditional probability of the ``person'' label appearing when the ``tie'' label is present. Given the proximity of the relationship between these two labels, P($l_\text{person}$|$l_\text{tie}$) tends to be high~\cite{zhao2021transformer, chen2021learning, ye2020attention}. 
  Suppose the ``person'' label originates from another task and remains unlabeled, while the ``tie'' is labeled. During training on the ``tie'', 
  the model tends to conflate the features of ``tie'' and ``person''. 
  Consequently, during the testing phase, even if only ``person'' is present in the image, the model tends to predict high values for both labels, resulting in an F-P error occurring on ``tie''. As more such label combinations are trained during class-incremental learning, the output distribution develops multiple peaks and becomes over-confident, leading to low uncertainty. 

 High confidence predictions correspond to output distributions with low uncertainty and vice versa \cite{pereyra2017regularizing, mukhoti2020calibrating}.
 Under the partial label setting, Shannon entropy serves as a well-defined metric to quantify uncertainty. Thus, we measure uncertainty for each old class in MLCIL using the following formula in task $t$:

\begin{equation}\label{eq:entropy}
H = - \mathbb E_{c \in \mc{C}^{1:t-1}} \left[\hat{y}_c^t \log(\hat{y}_c^t)\right] , 
\end{equation}
where $c$ denotes the class index, and $\hat{y}_c^t$  represents each confident score output by the model for task $t$. During the model optimization process, we negate the entropy $H$ to achieve maximum entropy regularization on the output distribution. The MLCIL model is calibrated by penalizing over-confident output distribution using the max-entropy regularization. 
\cref{fig:ent} (b) shows the statistics of confidence without or with max-entropy. Our method significantly reduces the F-P rate (FPR) from 35\% to 19\%. Moreover,  
the max-entropy also boosts the CF1, OF1, and mAP metrics shown in \cref{fig:ent} (a).
Note that the performance in experiments of either $c \in  \mc{C}^{1:t-1}$ or $c \in \mc{C}^{1:t}$ in \cref{eq:entropy} is similar, indicating that F-P errors predominantly occur within the old task classes. This is because past labels are often missing, catastrophic forgetting mainly occurs on old classes, where the old class features are more prone to confusion with each other. 



\subsection{Loss Function}

For the output, depicted in \cref{fig:CI-GCN}, we derive cross-task graph representations $\hat{y}^t_\text{gcn} \in \mathbb{R}^{|\mc{C}^{1:t}|}$ by passing $\textbf{V}_2$ through a fully connected layer, aiding in the final classification $\hat{y}^t$ by calibrating the label relationships.
Then, the multi-label confidence in task $t$ is calculated by:
\begin{equation}\label{eq:final}
\hat{y}^t = \hat{y}^t_\text{cls} + \hat{y}^t_\text{gcn},
\end{equation}
where class-related outputs $\hat{y}^t_\text{cls} \in \mathbb{R}^{|\mc{C}^{1:t}|}$ are obtained by passing class activation maps \textbf{M}  through global pooling. 
Similar to \cite{cermelli2020modeling,10221710,dong2023knowledge}, we combine cross-entropy with knowledge distillation (KD) to formulate our classification loss function: 
\begin{equation}\label{eq:cls}
L_{\text{cls}} = \underbrace{- \alpha \mathbb E_{c \in \mc{C}^{t}} \left[y_c^t \log(\hat{y}_c^t)\right]}_{\text{cross-entropy}} - \underbrace{(1-\alpha)\mathbb E_{c \in \mc{C}^{1:t-1}} \left[\hat{y}_c^{t-1} \log(\hat{y}_c^t)\right]}_{\text{knowledge distillation}} ,
\end{equation}
where $\alpha$ represents the loss weight, $y_c^t$ denotes the ground truth for $c \in \mc{C}^{t}$ and $\hat{y}_c^t$ signifies the prediction in task $t$. $\hat{y}_c^{t-1}$ represents the predictions from the previous task, utilized to preserve old task knowledge. Notably, $\hat{y}_c^{t-1}$ incorporates unlabeled supervisory information regarding old classes due to the task-level partial label issue, rendering it more effective in multi-label tasks than in single-label ones, as previously verified in \cite{cermelli2020modeling,10221710,dong2023knowledge}. Because $\textbf{A}_g$ can learn old task relationships from KD, it further enhances the effectiveness of KD in our method.


Combining the $H$ in \cref{eq:entropy}, our final loss function can be defined as:
\begin{equation}\label{eq:final}
L = L_{\text{cls}} - \beta H ,
\end{equation}
where $\beta$ is used to control the strength of max-entropy regularization.

\section{Experiments}
\subsection{Experimental Setting}
In this section, we provide a quantitative comparison between our approach and current state-of-the-art methods, along with comprehensive ablation studies. Additional results, including more max-entropy analysis, sensitive studies and visualizations, can be found in the supplementary material.

\noindent
\textbf{Datasets.} 
We follow the experimental setting of \cite{dong2023knowledge}, using MS-COCO 2014 \cite{lin2014microsoft} and PASCAL VOC 2007 \cite{everingham2010pascal} datasets to demonstrate the effectiveness of our method in MLCIL. MS-COCO serves as a widely used benchmark for multi-label classification. It consists of 82,081 images for training and 40,504 for
validation with 80 classes. The label average number for every image is 2.9. PASCAL VOC 2007  dataset comprises 20 classes. The training set contains 5011 images, the test set contains 4952 images.  On average, each image contains 2.4 labels.

Similar to these CIL works \cite{douillard2020podnet,dong2023knowledge,cha2021ssul}, we define different MLCIL scenarios using \{Bx-Cy\}, where ``x'' represents the number of classes trained in the base task, and ``y'' denotes the number of classes trained in each subsequent incremental task. We evaluate our method on MS-COCO with two different scenarios, \textit{i.e.}, \{B0-C10 and B40-C10\} as well as two more challenging scenarios, {\{B0-C5 and B20-C4\}}. Similarly, for VOC 2007, we evaluate our method with two scenarios, \textit{i.e.}, \{B0-C4 and B10-C2\} as well as two more challenging scenarios, {\{B5-C3 and B4-C2\}}. The incremental learning is the lexicographical order of class names, which is the same as \cite{dong2023knowledge}.

\noindent
\textbf{Evaluation Metrics.}
Similar to \cite{douillard2020podnet,dong2023knowledge}, we adopt the widely used average accuracy (Avg.Acc) and last accuracy (Last Acc) for CIL task. Following the MLCIL works \cite{10221710,dong2023knowledge}, we use the per-class F1 (CF1), overall F1 (OF1) as well as the mean average precision (mAP) to evaluate the MLCIL model. The mAP is divided into the average mAP and the last mAP, which means the average of the mAP of all tasks and the final task mAP, respectively.  During the testing phase, the label space for the last task is $\mc{Y}^T_\text{tst}=\mc{C}^{1:T}$, thus evaluation is conducted on the entire test set for the last task.

\noindent
\textbf{Implementation Details.}
For a fair comparison, we follow the experimental setting in \cite{dong2023knowledge} for all experiments. We adopt TResNetM \cite{ridnik2021tresnet} pre-trained on ImageNet-21k as the backbone. We train the model with a batch size of 64 for 20 epochs for MS-COCO 2014 and PASCAL VOC 2007. We optimize the network using Adam \cite{kingma2014adam} and OneCycleLR scheduler with a weight decay of 1e-4. We use an initial learning rate of 3e-5 for the base model and 5e-5 for the following tasks on MS-COCO, and use the learning rate of 4e-5 for all tasks on PASCAL VOC. We use the same data augmentation of \cite{dong2023knowledge}. 
\begin{table*}[t]
	\centering
	\caption{Class-incremental results on MS-COCO datasets. 
 Buffer size of 0 means no replay is applied and the * symbol denotes the results reported in \cite{dong2023knowledge}, same as below.} 
	\resizebox{\linewidth}{!}{
	\begin{tabular}{l|c|c|ccc|c|ccc|c}
		\hline
		\toprule
        \multirow{3}{*}{\textbf{Method}}& \multirow{2}{*}{\textbf{Method}}  & \multirow{2}{*}{\textbf{Buffer}}
        & \multicolumn{4}{|c|}{\textbf{MS-COCO B0-C10}} & \multicolumn{4}{|c}{\textbf{MS-COCO B40-C10}}  \\
        \cline{4-11}
		&\multirow{2}{*}{\textbf{Type}}&\multirow{2}{*}{\textbf{Size}} & \multicolumn{3}{c|}{Last Acc}& Avg.Acc & \multicolumn{3}{c|}{Last Acc}&  Avg.Acc  \\
        \cline{4-11}
		&&&mAP & CF1 & OF1& mAP  & mAP& CF1 & OF1 &  mAP  \\
        \hline
        Joint& Baseline  & \multirow{2}{*}{-} & 81.8 & 76.4 & 79.4 & - & 81.8 & 76.4 & 79.4 & -\\
        
        Fine-Tuning& Baseline  &  & 16.9 & 6.1 & 13.4& 38.3    & 17.0 & 6.0 & 13.6  & 35.1 \\
        
        \hline
        \hline
        oEWC \cite{schwarz2018progress}& SL  &\multirow{5}{*}{0}&   24.3 & 6.7 & 13.4 & 46.9& 27.3& 11.1 & 16.5& 44.8  \\
        LwF \cite{li2017learning}& SL  &&   42.4 & 45.3 & 43.7 & 61.2& 51.7& 47.0 & 45.7& 64.8  \\
        AGCN \cite{10221710}& ML  &  & 61.4 & 53.9 &  56.6 & 72.4  & 69.1& 58.7 & 59.9 & 73.9 \\
        
        KRT* \cite{dong2023knowledge}& ML  &   & 65.9& 55.6 & 56.5 & 74.6 & 74.0 & 64.4 & 63.4 & 77.8  \\
        \textbf{CSC}& ML  &   & \textbf{72.8}& \textbf{64.9} & \textbf{66.8}  & \textbf{78.0}   & \textbf{75.0}& \textbf{65.7} & \textbf{67.0}& \textbf{78.2}   \\
		\hline
  
        PODNet \cite{douillard2020podnet}& SL  & \multirow{5}{*}{5/class} & 53.4 & 13.6 & 17.3 & 65.7    & 57.8& 24.2 & 23.4& 65.4   \\
        DER++ \cite{buzzega2020dark}& SL  &  &   54.6 & 33.3 & 36.7& 68.1  &  59.0&  41.9&  43.7& 69.6 \\
        AGCN-R \cite{10221710}& ML  &   & 63.6& 56.7 & 58.5 & 72.9  &69.7 & 59.8 & 61.3 & 74.5 \\
        KRT-R* \cite{dong2023knowledge}& ML  &   & 68.3& 60.0 & 61.0 & 75.8   & 74.3 & 66.0 & 65.9 & 78.0  \\
        \textbf{CSC-R}& ML  &   &  \textbf{73.7}& \textbf{67.3} & \textbf{68.1}  & \textbf{79.2}  & \textbf{75.6}& \textbf{68.1} & \textbf{69.0}& \textbf{78.4}   \\
        \hline
        ER \cite{rolnick2019experience}& SL  & \multirow{6}{*}{20/class}  & 47.2 &40.6  & 43.6 & 60.3   & 61.6&58.6  & 61.1& 68.9   \\
        PODNet \cite{douillard2020podnet}& SL  &  & 58.8&45.2  &48.7 & 70.0     & 64.2 &  46.6& 42.1& 71.0  \\
        DER++ \cite{buzzega2020dark}& SL  &   & 63.1& 45.2 &48.7 & 72.7     & 66.3&51.5  & 53.5 & 73.6  \\
        AGCN-R \cite{10221710}& ML  &   & 66.0& 59.5 & 60.3 & 73.2  & 71.7& 64.1 & 65.2 & 75.2   \\
        KRT-R* \cite{dong2023knowledge}& ML  &  & 70.2 & 63.9 & 64.7& 76.5     & 75.2& 67.9 & 68.9& 78.3  \\
        \textbf{CSC-R}& ML  &  & \textbf{74.8}  &\textbf{67.8}  & \textbf{68.6}& \textbf{79.6}   & \textbf{76.7}& \textbf{68.2}  & \textbf{69.4} & \textbf{78.7}  \\
        \hline
        PRS\cite{kim2020imbalanced}& ML  & \multirow{5}{*}{1000} & 27.9&8.5  &14.7& 48.8    &33.2 & 9.3 & 15.1& 50.8   \\
        OCDM\cite{liang2022optimizing}& ML  &   &28.5& 8.6 &14.9& 49.5     &34.0 &9.5  &15.5  &51.3  \\
        AGCN-R \cite{10221710}& ML  &  & 59.0 & 59.4 & 65.9&73.0   & 71.1& 63.1 & 64.8& 75.0   \\
        KRT-R* \cite{dong2023knowledge}& ML  &   &69.3& 61.6 & 63.6 & 75.7   & 75.1& 67.5 & 68.5& 78.3  \\
\textbf{CSC-R}& ML  &  & \textbf{73.9} & \textbf{67.5}  & \textbf{68.5}   & \textbf{79.3} &\textbf{ 76.0}& \textbf{67.8} & \textbf{69.7}& \textbf{78.5}    \\
		\bottomrule
	\end{tabular}}
\label{tab:results_1}
\end{table*}

\subsection{Comparison Methods}
We select several advanced methods for comprehensive comparison.

\noindent
\textbf{Baseline Methods.}
Following  KRT \cite{dong2023knowledge} and AGCN \cite{10221710}, we select Fine-Tuning and Joint as baseline methods. Fine-Tuning performs training without any anti-forgetting technique. It can be regarded as the performance lower bound. Joint is the upper bound, which trains the model jointly using all task data.

\noindent
\textbf{Single-Label Class-Incremental Methods.}
We run six single-label (SL) class-incremental learning methods with our MLCIL setting, including the most representative regularization-based methods oEWC \cite{schwarz2018progress} and LwF \cite{li2017learning}, and the widely recognized replay-based methods ER \cite{rolnick2019experience}, PODNet \cite{douillard2020podnet} and DER++ \cite{buzzega2020dark}. We also compare with the recent SLCIL prompt-based method L2P \cite{wang2022learning} based on ViT-B/16 backbone, to show the gap from the respective upper bound.

\noindent
\textbf{Multi-Label Class-Incremental Methods.}
We select four best-performing multi-label (ML) class-incremental methods to compare, including replay-based methods PRS \cite{kim2020imbalanced} and OCDM \cite{liang2022optimizing}, the regularization-based methods AGCN \cite{10221710} and KRT  \cite{dong2023knowledge}. Note that KRT  \cite{dong2023knowledge} is the current SOTA. We compare the results with those reported in KRT  \cite{dong2023knowledge} under strictly the same experimental setting.

\noindent
\textbf{Our Method.}
CSC is our method without a replay buffer, and CSC-R is CSC with the replay buffer for a fair comparison with replay-based methods.


\subsection{Comparison Results}
\noindent
\textbf{MS-COCO.}  The results in \{B0-C10 and B40-C10\} of MS-COCO are shown in \cref{tab:results_1}.  CSC outperforms other advanced methods in all scenarios on MS-COCO. We have the following observations: 1) CSC achieves the best final mAP of \textbf{74.8\%} and \textbf{76.7\%} when the buffer size is 20/class. 2) When the buffer size gets smaller (5/class), our method outperforms the SOTA method KRT more: the improvements on final mAP, CF1, and OF1 are \textbf{5.4\%} (68.3\%$\rightarrow$73.7\%), \textbf{7.3\%} (60.0\%$\rightarrow$67.3\%), and \textbf{7.1\%}  (61.0\%$\rightarrow$68.1\%) in \{B0-C10\} scenario; \textbf{1.3\%}  (74.3\%$\rightarrow$75.6\%), \textbf{2.1\%} (66.0\%$\rightarrow$68.1\%) and \textbf{3.1\%}  (65.9\%$\rightarrow$69.0\%) in \{B40-C10\} scenario. 3) It is evident that \{B0-C10\} scenario is more challenging than \{B40-C10\}. In more challenging scenarios, our method is superior to others. When the buffer size is set to 0, our method achieves a more significant leap in performance,  surpassing the current SOTA by a wide margin, improved by \textbf{6.9\%} (65.9\%$\rightarrow$72.8\%), \textbf{9.3\%} (55.6\%$\rightarrow$64.9\%), and \textbf{10.2\%} (56.6\%$\rightarrow$66.8\%) on the mAP, CF1, and OF1 in \{B0-C10\}, respectively. And 4) compared to KRT-R (20/class, \textbf{70.2\%} in mAP), CSC without replay buffer (0, \textbf{72.8\%} in mAP) also achieves superior results in the more challenging scenario \{B0-C10\}.

\begin{table*}[t]
	\centering
	\caption{Class-incremental results on PASCAL VOC datasets. 
 } 
	\resizebox{\linewidth}{!}{
	\begin{tabular}{l|c|cc|cc|cc|cc}
		\hline
		\toprule
        \multirow{2}{*}{\textbf{Method}} & \textbf{Buffer}
        & \multicolumn{2}{|c|}{\textbf{VOC B0-C4}} & \multicolumn{2}{|c}{\textbf{VOC B10-C2}} & \multicolumn{2}{|c}{\textbf{VOC B5-C3}} & \multicolumn{2}{|c}{\textbf{VOC B4-C2}}\\
        \cline{3-10}
		&\textbf{Size}& Last mAP& Avg.mAP  & Last mAP&  Avg.mAP  & Last mAP&  Avg.mAP   & Last mAP&  Avg.mAP   \\

        \hline
        Joint & \multirow{2}{*}{-}& 93.6&-  & 93.6& - & 93.6& - & 93.6& - \\
        Fine-Tuning  & & 62.9& 82.1  & 43.0& 70.1  &49.4&74.4&37.0&60.4 \\
        \hline
        \hline
        AGCN \cite{10221710}  & \multirow{2}{*}{0} & 73.4& 84.3 & 65.1 & 79.4   & 69.2& 80.9  & 53.4& 72.1  \\
        \textbf{CSC}  &  & \textbf{85.1}& {\textbf{90.4}}  & \textbf{83.8}& \textbf{89.0}  &  \textbf{82.1}&\textbf{88.0}  & \textbf{74.1}& \textbf{83.3}  \\
        \hline
        ER\cite{rolnick2019experience} &  \multirow{6}{*}{2/class}& 67.7& 82.6  &64.3&78.8  &65.3&79.0  &57.9&73.7 \\
        PODNet \cite{douillard2020podnet}   &  & 72.3& 84.1 & 66.2 & 80.2  & 70.3& 81.4 &62.5 & 75.7\\
        DER++\cite{buzzega2020dark} & &74.4& 85.0 &66.2&80.7&68.1&78.0&61.6&77.0\\
        
        AGCN-R \cite{10221710}  &  & 76.0& 86.5  & 69.3& 82.8  & 71.8& 82.1  &59.3& 74.3 \\
        KRT-R* \cite{dong2023knowledge}  &  & 83.4& 90.7 & 80.5 & 87.7  & - & - & - &- \\
        \textbf{CSC-R}  &  & \textbf{87.9}& \textbf{92.4}  & \textbf{87.8}& \textbf{91.6} & \textbf{87.5} & \textbf{91.9}  & \textbf{86.6}& \textbf{90.4}  \\
		\bottomrule
	\end{tabular}}
\label{tab:results_3}
\end{table*}
\begin{table*}[t]
        \centering
        \caption{Comparative experimental results under different backbones.
        }
  	\resizebox{0.8\linewidth}{!}{
	\begin{tabular}{l|c|c|ccc|c|ccc|c}
		\hline
		\toprule
        \multirow{3}{*}{\textbf{Method}}& \multirow{3}{*}{\textbf{Backbone}}  & \multirow{3}{*}{\textbf{Param.}}
        & \multicolumn{4}{|c|}{\textbf{MS-COCO B0-C10}} & \multicolumn{4}{|c}{\textbf{MS-COCO B40-C10}}  \\
        \cline{4-11}
		&& & \multicolumn{3}{c|}{Last Acc} & Avg.Acc & \multicolumn{3}{c|}{Last Acc}&  Avg.Acc \\
        \cline{4-11}
		&&& mAP& CF1 & OF1 & mAP & mAP & CF1 & OF1 & mAP  \\
        \hline
        Joint&\multirow{2}{*}{ViT-B/16}  & \multirow{2}{*}{86.0M} & 83.2 & 78.3 & 80.5 & - & 83.2 & 78.3 & 80.5 &-\\
        L2P \cite{wang2022learning} & & &68.0 & 58.8 & 56.0&73.4 &71.1&62.8&62.3&73.7 \\
        \hline
        Joint & \multirow{3}{*}{TResNetM}&\multirow{3}{*}{29.4M} &81.8 &76.4&79.4& - & 81.8&76.4&79.4& - \\
        KRT* \cite{dong2023knowledge}&  &  &65.9 & 55.6 & 56.5 & 74.6  & 74.0 & 64.4 & 63.4 &77.8  \\
        \textbf{CSC}&  &  & \textbf{72.8} & \textbf{64.9} & \textbf{66.8} & \textbf{78.0}   & \textbf{75.0} & \textbf{65.7} & \textbf{67.0} & \textbf{78.2}   \\
        \bottomrule
           
        \end{tabular}}
        
\label{tab:results_2}
\end{table*}

The results on the two more challenging scenarios \{B0-C5 and B20-C4\} of MS-COCO are shown in \cref{fig:figure3}. These mAP curves demonstrate the significant superiority of our method during the CIL process. Moreover, our method exhibits a closer proximity to the upper bound. This demonstrates that our method performs effectively even in long-term class-incremental processes. 
\begin{table*}[t]
        \centering
        \caption{The ablation study for effectiveness of each component.
        }
  	\resizebox{0.9\linewidth}{!}{
	\begin{tabular}{l|cc|ccc|c|ccc|c}
		\hline
		\toprule
        \multirow{2}{*}{\textbf{Method}}
        &\multirow{2}{*}{\textbf{Max-Entropy}}&\multirow{2}{*}{\textbf{CI-GCN}}& \multicolumn{4}{|c|}{\textbf{MS-COCO B0-C10}} & \multicolumn{4}{|c}{\textbf{MS-COCO B0-C5}}  \\
        \cline{4-11}
		
        \cline{4-11}
		&&&mAP&CF1 & OF1 &Avg. mAP & mAP & CF1 & OF1 & Avg. mAP  \\
        \hline
Baseline&&&42.4&45.3&43.7&61.2&35.9&30.0&28.0&54.9\\
        \hline
        (a) CSC &\checkmark&&{47.6}&{50.3}&{49.5}&63.1&39.2&35.4&33.1&57.3\\
        (b) CSC&&\checkmark&69.3&59.0&59.5&75.6&64.8&42.6&39.5&73.4\\
        
        (c) CSC&\checkmark&\checkmark&\textbf{72.8}&\textbf{64.9}&\textbf{66.8}&\textbf{78.0}&\textbf{67.3}&\textbf{50.5}&\textbf{47.2}&\textbf{75.0}\\
        \bottomrule
           
        \end{tabular}}
        
\label{tab:results_4}
\end{table*}
\begin{figure*}[t]
  \begin{minipage}{0.42\linewidth}
    \centering
    \captionof{table}{Correlation matrix study.  ``Z'': fixed statistical CM, ``G'': general CM and ``S'': specific CM.}
    \resizebox{\linewidth}{!}{
\begin{tabular}{l|ccc|c}
		\hline
		\toprule
        \multirow{2}{*}{\textbf{Structure}}
		&  \multicolumn{3}{c|}{Last Acc} & Avg.Acc \\
        \cline{2-5}
		& mAP& CF1 & OF1 & mAP  \\
        \hline
        Z $\rightarrow$ Z (sigmoid) & 64.1 & 57.1 & 59.1 & 73.7 \\
        Z $\rightarrow$ S (sigmoid) & 69.8 & 62.6 & 65.3 &75.6  \\
        G $\rightarrow$ S (sigmoid) &  71.2  & 64.0 & 65.5 & 76.5 \\
        G $\rightarrow$ S (softmax)  & \textbf{72.8}& \textbf{64.9} &\textbf{66.8} &\textbf{78.0} \\

        \bottomrule
        \end{tabular}
    }

    \label{tab:results_5}
  \end{minipage}
  \hfill
  \begin{minipage}{0.54\linewidth}
    \centering
    \captionof{table}{Sensitive study of max-entropy weight $\beta$ and buffer size from 5/class to 50/class in \{B0-C10\} of MS-COCO.}
    \resizebox{\linewidth}{!}{
        	\begin{tabular}{l|ccc|c|ccc}
		\hline
		\toprule

		$\beta$&mAP& CF1 & OF1&Buffer Size&mAP& CF1 & OF1  \\
        \hline
        1e-3 & 72.97& 62.54&62.92&5/class&73.68&67.33&68.09 \\
            2e-3 &\textbf{73.02} &63.84 &64.24&10/class&73.93&67.54&68.53 \\
            4e-3 &72.84 &\textbf{64.89} &\textbf{66.83}&20/class&74.78&67.82&68.56 \\
            6e-3 &71.71 &63.57 &66.66&30/class&75.01&67.89& 68.90\\
            8e-3 &71.13 &60.56 &65.53&50/class&\textbf{75.26}&\textbf{68.08}&\textbf{68.97} \\
        \bottomrule 
        \end{tabular}
    }

    \label{tab:results_6}
  \end{minipage}
\end{figure*}
\noindent
\textbf{PASCAL VOC.}  The results in \{B0-C4, B10-C2, B5-C3 and B4-C2\} of PASCAL VOC are shown in \cref{tab:results_3}. Just as on MS-COCO, our method significantly outperforms the state-of-the-art in all scenarios. Specifically, the final mAP improved by\textbf{ 4.5\%} (83.4\%$\rightarrow$87.9\%) and \textbf{7.3\%} (80.5\%$\rightarrow$87.8\%) in \{B0-C4\} and \{B10-C2\}, respectively. Furthermore, our method exhibits better robustness across different scenarios. For instance, from \{B0-C4\} to \{B4-C2\}, the final mAP of CSC-R changes from 87.9\% to 86.6\% ($\downarrow$ \textbf{1.3\%}). However,  the final mAP of AGCN-R changes from 76.0\% to 59.3\% ($\downarrow$ \textbf{16.7\%}). This demonstrates that, owing to CI-GCN and max-entropy regularization, our method possesses better environmental adaptability and robustness.
The every task result in \{B5-C3 and B4-C2\}  scenarios of PASCAL VOC is shown in \cref{fig:figure3}. Consistent with the conclusions drawn from the MS-COCO, our method outperforms other comparative methods across all class-incremental tasks.

\noindent
\textbf{Comparison of different backbones.}
\cref{tab:results_2} illustrates the comparison between our method and the prompt-based method L2P \cite{wang2022learning}. The backbone of L2P is the pre-trained ViT-B/16. Our method and KRT \cite{dong2023knowledge} are based on the pre-trained TResNetM. 
It can be observed that L2P performs better overall compared to KRT in the more challenging \{B0-C10\} scenario.
However, our method achieves better results with fewer parameters than L2P in both scenarios.


           
        
\begin{figure}[t]
    \centering
    \includegraphics[width=\linewidth]{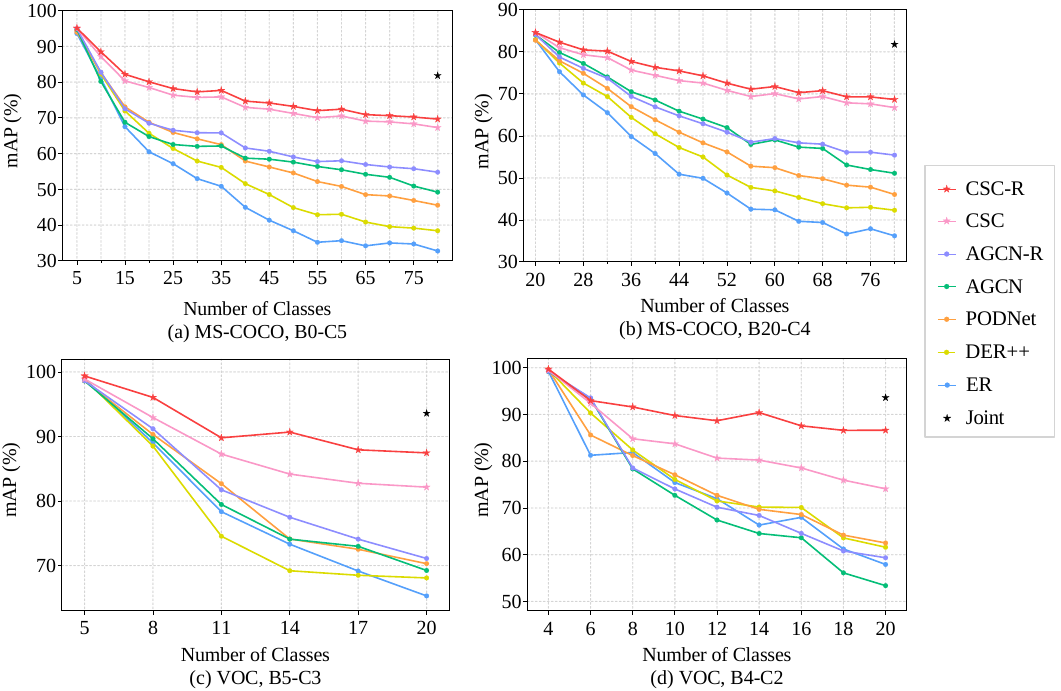}
    \caption{Comparison results on MS-COCO and VOC datasets in challenging scenarios. There are more incremental tasks in these scenarios.} 
    \label{fig:figure3}
\end{figure}
\subsection{Ablation Study} 
\noindent
\textbf{Effectiveness of Each Component.}  \cref{tab:results_4} shows the detailed analysis of our contributions in two challenging scenarios. Under the task-level partial label setting, the knowledge distillation provides supervised information for missing labels, thus we naturally utilize it as a baseline, similar to \cite{cermelli2020modeling,10221710,dong2023knowledge}. 
Firstly, when we calibrate the baseline model confidence with max-entropy (row a), the mAP, CF1 and OF1 of the model are significantly improved. Then, when the disjoint label spaces are connected by constructing cross-task relationships (row b), the improvement is remarkable. For example, the mAP, CF1 and OF1 of the model in \{B0-C10\} are improved by \textbf{26.9\%} (42.4\%$\rightarrow$69.3\%), \textbf{13.7\%} (45.3\%$\rightarrow$59.0\%) and \textbf{15.8\%} (43.7\%$\rightarrow$59.5\%). Finally, we add the max-entropy regularization to the model (row c),  this component can further improve the model's performance by calibrating the multi-label confidence: \textbf{64.8\%} vs \textbf{67.3\%} in mAP,  \textbf{42.6\%} vs \textbf{50.5\%} in CF1 and \textbf{39.5\%} vs \textbf{47.2\%} in OF1 of \{B0-C5\}. Furthermore, in horizontal comparisons, we found that even in more challenging scenarios, our approach still demonstrates stable superiority \textbf{72.8\%} vs \textbf{67.3\%} in mAP. These results demonstrate that CSC framework can significantly suppress the negative impact of partial label and make multi-label confidence well-calibrated.

           
        
\begin{figure}[t]
    \centering
    \includegraphics[width=\linewidth]{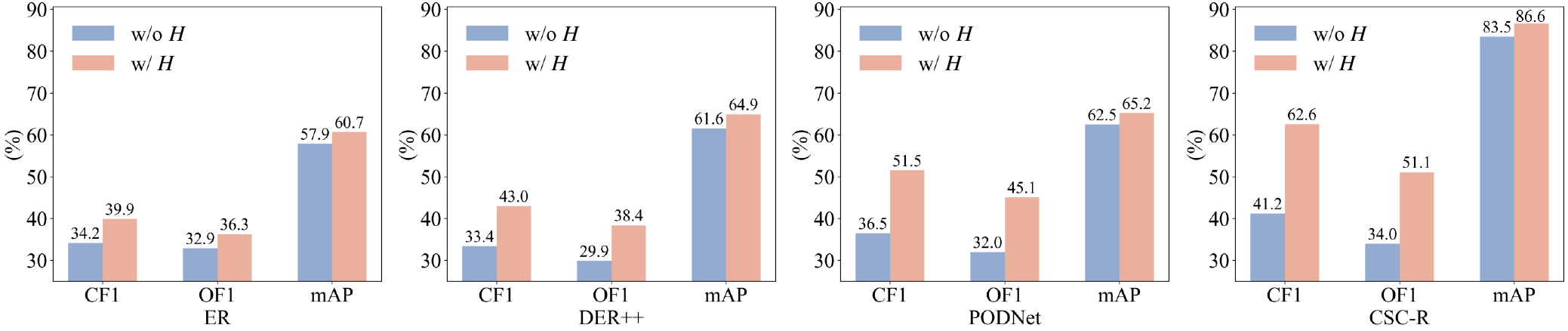}
    \caption{Max-entropy analysis in \{B4-C2\} of VOC.} 
    \label{fig:figure5}
\end{figure}
\begin{figure*}
    \centering
    \includegraphics[width=\linewidth]{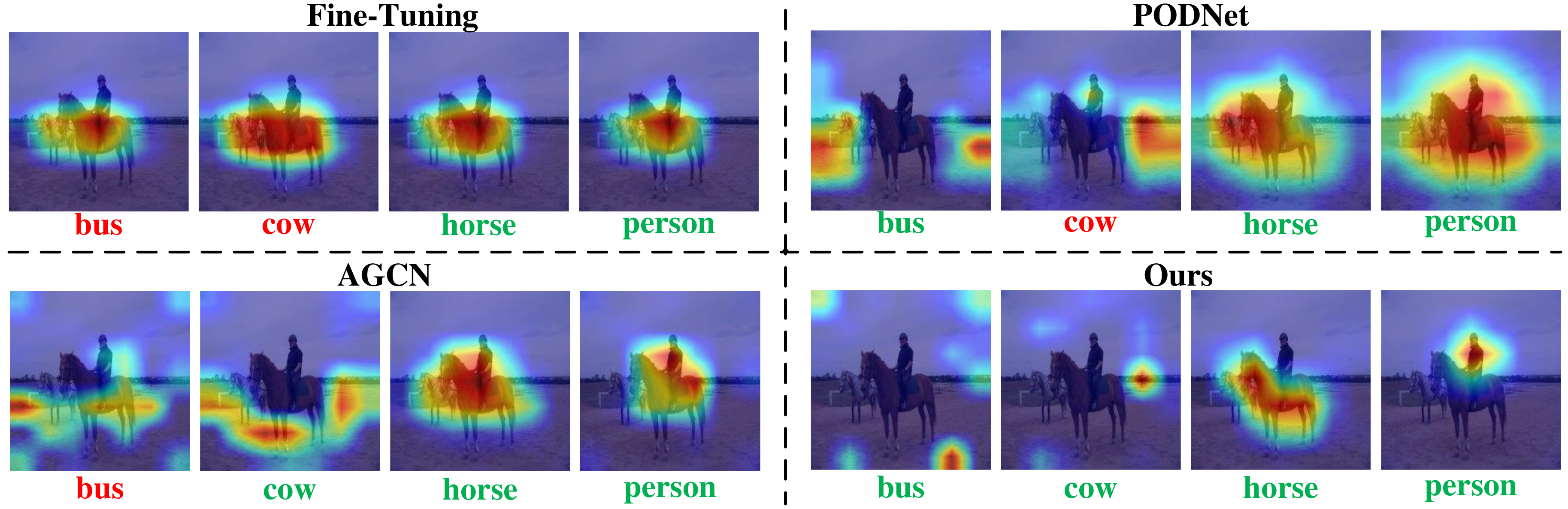}
    \caption{Visualization of CIL methods. Labels in red: F-P error, green: correctness. 
    }
    \label{fig:Visualization}
\end{figure*}
\noindent
\textbf{Correlation Matrix Study.}   
To verify the effectiveness of the general and specific CM, we show the performances of different combinations of CM in \cref{tab:results_5}. ``Z'' denotes the CM constructed using a statistical way like \cite{10221710,chen2019multi}, which is fixed for each sample. ``G'' means our general CM  and ``S'' means specific CM. \cref{tab:results_5} also shows ablation studies on regularization $Sigmoid(\cdot)$ and $Softmax(\cdot)$ functions, which are used for regularization of class activation maps $\textbf{M}$ in \cref{eq:v}. When the combination way is a general CM followed by a specific CM (G $\rightarrow$ S) with $Softmax(\cdot)$, the results are superior. The results show that constructing class-incremental label relationships in CI-GCN is more effective than fixed statistical label relationships.

\noindent
\textbf{Sensitive Study.}   
As shown in \cref{tab:results_6}, to further demonstrate the robustness of our approach, we conduct the sensitive study. We first set $\alpha$ in \cref{eq:cls} to 0.15. Then, we analyze the influence of $\beta$ in \cref{eq:final}, when the $\beta$ is set to 4e-3, the overall performance of the model is superior. We increase the buffer size from 5/class to 50/class, the final mAP increases from 73.68\% to 75.26\%. Furthermore, when the buffer size is set to 0, the final mAP is 72.84\%.
These results indicate the robustness of our method, demonstrating its capability to effectively preserve knowledge and mitigate forgetting even without a buffer.

\noindent
\textbf{Max-Entropy Analysis.} 
\cref{fig:figure5} demonstrates the orthogonality of max-entropy calibration across various class-incremental methods under different evaluation metrics. These results provide evidence that max-entropy regularization can significantly enhance the performance of different methods.
\subsection{Visualization} 
\cref{fig:Visualization} shows several attention maps for four methods. The labels ``bus'', ``cow'', ``horse'' and ``person'' are sequentially trained in different tasks. After training on ``person'', given a test image with ``horse'' and  ``person'', the model outputs attention maps for the four classes. 
It can be seen that our calibrated model can output appropriate confidence to achieve correct predictions and accurately capture the features of each class. 

\section{Conclusion}
In this paper, we studied the less-explored task of multi-label class-incremental learning (MLCIL). 
 In MLCIL, only the new classes are labeled at training time, the absence of past and future labels results in a \textit{task-level partial label} issue for the current task. In this paper, we refined the multi-label confidence calibration in MLCIL and proposed a Confidence Self-Calibration (CSC) framework for addressing this issue. Firstly, we introduced a class-incremental graph convolutional network (CI-GCN) to calibrate the label relationships by bridging the isolated label spaces. 
 Then, we presented a max-entropy regularization for over-confident output distribution to facilitate confidence self-calibration. 
Our approach achieved new SOTA results in MLCIL tasks on both MS-COCO and PASCAL VOC datasets.
\clearpage


%
%
\bibliographystyle{splncs04}
\bibliography{egbib}
\end{document}